\documentclass{ecai} 

\usepackage{latexsym}
\usepackage{amssymb}
\usepackage{amsmath}
\usepackage{amsthm}
\usepackage{booktabs}
\usepackage{enumitem}
\usepackage{graphicx}
\usepackage{color}
\usepackage{times}
\usepackage{soul}
\usepackage{url}
\usepackage[hidelinks]{hyperref}
\usepackage[utf8]{inputenc}
\usepackage[small]{caption}
\usepackage{graphicx}
\usepackage{amsmath}
\usepackage{amsthm}
\usepackage{booktabs}
\usepackage[switch]{lineno}
\usepackage[ruled,vlined]{algorithm2e}

\usepackage{graphicx}
\usepackage[export]{adjustbox} 

\usepackage{enumitem}

\usepackage{comment}
\usepackage{amssymb}
\usepackage{amsmath}
\usepackage{amsmath}
\usepackage{mathtools}
\usepackage{xcolor}
\usepackage{stmaryrd}

\usepackage{comment}
\usepackage{amsfonts}
\usepackage{dsfont}
\usepackage{array}

\usepackage{multirow}
\usepackage{subfigure}
\usepackage{caption}
\usepackage{bbding}
\usepackage{amssymb}
\newcommand{\SH}[1]{\textcolor{blue}{#1}}

\newcommand{\BibTeX}{B\kern-.05em{\sc i\kern-.025em b}\kern-.08em\TeX}

\begin{document}


\begin{frontmatter}


\paperid{123} 


\title{NAEx: A Plug-and-Play Framework for \\ Explaining Network Alignment}

\author[A]{\fnms{Shruti}~\snm{Saxena}\thanks{Corresponding Author. Email: shruti\_2021cs31@iitp.ac.in}}
\author[B]{\fnms{Ariijt}~\snm{Khan}}
\author[A]{\fnms{Joydeep}~\snm{Chandra}} 

\address[A]{Indian Institute of Technology Patna, India}
\address[B]{Aalborg University, Denmark}

\begin{abstract}
Network alignment (NA) identifies corresponding nodes across multiple networks, with applications in domains like social networks, co-authorship, and biology. Despite advances in alignment models, their interpretability remains limited, making it difficult to understand alignment decisions and posing challenges in building trust, particularly in high-stakes domains. To address this, we introduce \textsf{NAEx}, a plug-and-play, model-agnostic framework that explains alignment models by identifying key subgraphs and features influencing predictions. \textsf{NAEx} addresses the key challenge of preserving the joint cross-network dependencies on alignment decisions by: (1) jointly parameterizing graph structures and feature spaces through learnable edge and feature masks, and (2) introducing an optimization objective that ensures explanations are both faithful to the original predictions and enable meaningful comparisons of structural and feature-based similarities between networks. \textsf{NAEx} is an inductive framework that efficiently generates NA explanations for previously unseen data. We introduce evaluation metrics tailored to alignment explainability and demonstrate \textsf{NAEx}'s effectiveness and efficiency on benchmark datasets by integrating it with four representative NA models.

\end{abstract}

\end{frontmatter}

\section{Introduction}\label{sec:intro}
Network Alignment (NA) is the task of finding node correspondences across networks \cite{trung2020comparative}. It is widely studied for integrating and analyzing information from multiple co-existing networks within the same domain, such as social networks, e.g., {\em Facebook}-{\em Instagram}, co-authorship networks, e.g., {\em AMiner}-{\em DBLP}, biological networks, e.g., {\em yeast} and {\em human PPI networks}, information networks, e.g., {\em Wikidata} and {\em IMDB}, etc. NA is considered a pre-requisite for various inter-network applications, including cross-site friends recommendation, products recommendation, fraud detection, and uncovering novel interaction patterns in protein networks \cite{LIU2018318,btr127,LiuZP17,CARMAGNOLA200916,ConteFSV03}.

Recent network alignment methods have been developed on the basis of two underlying assumptions- \textit{attribute consistency}- the attributes of mapped nodes are similar, and \textit{structural consistency}- the mapped nodes exhibit similar neighborhood structure \cite{trung2020adaptive,ZRJZZJYD21}. Harnessing the high expressive power of graph neural networks (GNNs), recent advancements predominantly utilize GNNs \cite{trung2020adaptive,liang2021unsupervised,huynh2021network,zheng2022jora,park2022gradalign+,sun2023towards,li2025attention,seo2024leveraging}, which generates node embeddings by aggregating the attributes of its neighbors. When optimized over a joint learning framework, these node embeddings inherently capture both attribute and structural consistencies to effectively align networks \cite{trung2020adaptive}. While these works on NA have achieved significant accuracy improvements, their interpretability remains limited.
They rely solely on numerical evaluation metrics without providing insights into the relationships or attributes that derive the alignment decisions. A good explanation model would help understand  ``why'' a particular NA model aligns a node pair (e.g., Alice on {\em Facebook} is aligned to Alicia on {\em LinkedIn} primarily because of the shared location, profession, and mutual connections from the same workplace). Understanding NA models can also play a crucial role in distinguishing them and highlighting their assumptions, strengths, and weaknesses, thereby aiding task-specific model selection. Additionally, explainable models foster trust and adoption in real-world applications where decision-makers often require interpretability \cite{saxena2024survey}.

\par\noindent\textbf{Challenges.} The limited efforts in explaining NA methods pose the first challenge.
A study by Zhou et al. \cite{zhou2019disentangled} makes an effort to provide explainability using the idea of influence functions, giving insights into the influence of individual nodes (positive or negative) on a particular prediction. However, it has several limitations: (1) It evaluates nodes in isolation and does not account for the combined influence of multiple nodes, edges, and node attributes, which can better represent the alignment context. (2) Approximating the effect of each node on the model parameters is computationally intensive. (3) It requires parameter re-evaluation for each node and prediction, limiting its inductive capability to explain new predictions.

Similarly, efforts have been made for the recent advancements in the explainability of GNNs \cite{YuanYGJ23}. The explanations are in the form of the most influential subgraph structure and/or a set of node features crucial for the GNN's prediction on the target instance (a node or a graph). However, applying these methods directly to NA tasks is challenging. The classical objective function of explaining GNNs is based on mutual information between the GNN's prediction and an explanation subgraph \cite{YingBYZL19} -- this is defined at a single graph level and does not directly handle the multi-network alignment challenges. Explaining an alignment output (e.g., Alice $\equiv$ Alicia) requires modelling cross-network dependencies,
identifying key substructures, attributes, or relationships that justify their alignment. Additionally, the explanation generated from multiple networks must be consistent for direct comparison of their structural and feature-based similarities.

\par\noindent\textbf{Solution.} To address the above challenges, we propose \textsf{NAEx}, a post hoc and model-agnostic explanation framework specifically designed for neural embedding-based NA models. The novelty of our proposed approach lies in a mutual information-based  and general-purpose framework that jointly learns masked subgraph pairs around anchor nodes across source and target networks. In particular, we extend the GNN-based explanations to multi-network settings along with overcoming the challenges of the existing NA explanation method. \textsf{NAEx} introduces a generative probabilistic approach that models the graph structures as edge distributions and generates explanatory subgraphs that reveal why an NA model aligns a specific pair of nodes from the two networks. The explanation generation process is parameterized using a deep neural network, enabling the framework to explain multiple alignment predictions collectively. The parameters of this explanation network are shared across the networks, and all alignment instances, thereby supporting a global understanding of the NA model. \textsf{NAEx} also supports feature explanation by recognizing the set of most important features for the alignment prediction. We introduce two objectives to optimize \textsf{NAEx}: an alignment consistency objective that ensures the prediction fidelity of the model and a subgraph similarity objective that promotes the correspondence between the generated explanations for their direct interpretation. A key strength of \textsf{NAEx} is its ability to generalize. Once trained, it can infer explanations for unexplained node pairs in an inductive setting without retraining, making it a highly efficient plug-and-play framework. We evaluate \textsf{NAEx} by generating explanations for four state-of-the-art NA model predictions over three real-world datasets from diverse domains. We introduce three evaluation metrics, Fidelity, Faithfulness, and Sparsity, specially tailored to the alignment task without the need for ground truth explanations.




\par\noindent\textbf{Contributions.} We summarize our key contributions below.
\begin{enumerate}
    \item We propose \textsf{NAEx}, the first model-agnostic, post hoc, and general-purpose framework for explaining neural embedding-based NA predictions that work on both transductive and inductive settings. 
    \item We formally define a parameterized objective function for explaining the NA predictions and introduce an optimization framework that ensures prediction fidelity while preserving structural and semantic consistency in the generated explanations.
    \item We introduce evaluation metrics to assess the quality of explanations specific to the alignment task without the need for ground truth explanations.
    \item We apply \textsf{NAEx} on four representative NA methods over three benchmark datasets. \textsf{NAEx} delivers $\sim 32\%$ improvement in explanation fidelity and achieves up to $95\%$ speedup compared to the perturbation-based baseline dNAME \cite{zhou2019disentangled}.
    
\end{enumerate}




\section{Related Work}
NA approaches \cite{saxena2024survey,TANG20251} are broadly categorized into matrix factorization-based \cite{koutra2013big,zhang2016final,heimann2018regal} and embedding-based \cite{YanLLCZLW23,trung2020adaptive,liang2021unsupervised} methods. 
Recent state-of-the-art embedding based methods leverage graph neural networks (GNNs) for generating node representations and employ end-to-end frameworks for joint embedding and alignment learning \cite{trung2020adaptive,liang2021unsupervised,huynh2021network,zheng2022jora,park2022gradalign+,sun2023towards,li2025attention,seo2024leveraging}. 

Explainability techniques for GNNs are categorized into intrinsic and post hoc approaches \cite{kakkad2023survey,khan2023interpretability}. Intrinsic methods, like Graph Attention Networks (GATs) \cite{velickovic2017graph}, build explainability into the model, while post hoc methods generate separate models for explanations \cite{YingBYZL19}. Techniques are also classified as global (focusing on model behavior) \cite{yuan2020xgnn,luo2020pgexplainer} or local (explaining individual predictions) \cite{YingBYZL19}.

Few NA explanation frameworks exist. \cite{zhou2019disentangled} is a perturbation-based, local method that assesses each node's influence on a prediction but fails to capture node relationships, is inefficient, and lacks inductive capability. Other methods focus on rule-based explanations for knowledge graphs \cite{tian2024generating,yeo2018xina} and rely heavily on edge relations.
 \textsf{NAEx} is the first general-purpose, model-agnostic, post hoc, and global method for explaining GNN-based NA models, as detailed in Table \ref{tab:rel_works}. More details are provided in the Supplementary.

\begin{table}[t]
\caption{ \small Characteristics of NA explanation methods. \checkmark if the method possesses the characteristic, otherwise \XSolidBrush. An ideal approach should be model agostic, post hoc, inductive and should not be KG specific. Our proposed \textsf{NAEx} method fulfills all these criteria.}
\centering
\label{tab:my-table}
\setlength{\extrarowheight}{2pt}
\resizebox{\columnwidth}{!}{%
\begin{tabular}{l|c|c|c|c|c}
\hline
\multicolumn{1}{c|}{} & \begin{tabular}[c]{@{}c@{}}dNAME\\ \cite{zhou2019disentangled}\end{tabular}& \begin{tabular}[c]{@{}c@{}}XINA\\ \cite{yeo2018xina}\end{tabular}& \begin{tabular}[c]{@{}c@{}}ExEA\\ \cite{tian2024generating}\end{tabular}& \begin{tabular}[c]{@{}c@{}}i-Align\\ \cite{trisedya2023align}\end{tabular}& \begin{tabular}[c]{@{}c@{}}NAEx\\ (Ours)\end{tabular} \\ \hline
Model agnostic & \checkmark   & \XSolidBrush & \checkmark   & \XSolidBrush & \checkmark   \\
Post hoc       & \checkmark   & \XSolidBrush & \checkmark   & \XSolidBrush   & \checkmark   \\
Transductive      & \checkmark & \checkmark & \checkmark & \checkmark & \checkmark   \\ 
Inductive      & \XSolidBrush & \XSolidBrush & \XSolidBrush & \XSolidBrush & \checkmark   \\ 
Heterogeneous graph (e.g. KG)    & \checkmark & \checkmark   & \checkmark   & \checkmark & \checkmark \\
Homogeneous graph    & \checkmark & \XSolidBrush   & \XSolidBrush   & \XSolidBrush & \checkmark\\ \hline
\end{tabular}%
}
\label{tab:rel_works}
\end{table}

\section{Preliminaries}
We provide background on embedding-based network alignment and the task of explaining graph neural networks.

\subsection{Embedding-based Network Alignment}
Let $G=(V,E,X)$ denote a graph with the node set $V$, edge set $E\subseteq V\times V$, and node attribute matrix $X\in \mathbb{R}^{|V| \times d_m}$. Given a source network $G_s=(V_s, E_s, X_s)$ and a target network $G_t=(V_t, E_t, X_t)$, the goal of embedding-based network alignment is to learn a mapping $\pi: V_s \rightarrow  V_t$ that associates nodes from the source network to the target network such that they represent the same entity in two different networks. To achieve this, a model  $F_\theta$ is trained to learn node embeddings $ h^s_u = F_\theta(u \mid G_s)$, and $h^t_v = F_\theta(v \mid G_t)$, where where \( h^s_u, h^t_v \in \mathbb{R}^{d_h} \). These embeddings aim to capture both structural and attribute-based information and are designed such that the mapped nodes have similar representations in the embedding space. The alignment is performed by the function comparing embeddings:
\begin{equation}
    \pi(u) = \arg\max_{v \in V_t} \text{sim}(h^s_u, h^t_v), \quad \forall u\in V_s
\end{equation}
where \( \text{sim}(\cdot, \cdot) \) is a similarity function such as cosine similarity, dot product, or Euclidean distance. Then, the predicted alignment set \( A \) is given by:
\(A = \{(u, \pi(u)) \mid u \in V_s\}\). In practical scenarios, a mutual consistency is required for which a  reverse alignment function \( \pi': V_t \to V_s \) is used and the final alignment set includes only symmetric matches:
\(A = \{(u, v) \mid \pi(u) = v \text{ and } \pi'(v) = u\}.\)
In a weakly supervised setting, the model $F_\theta$ is trained using a small set of pre-aligned node pairs $A_{\text{train}}$~\cite{tangWeaklySupervisedWSDM23}. Such models are typically transductive, meaning that they are designed to generate node mappings only for the specific networks on which they are trained.

\subsection{Explanation for network alignment} 
We aim to provide post hoc explanations for specific predictions made by NA models. Given an aligned pair $(u \equiv v)$ predicted by a trained NA model  $F_\theta$, our goal is to identify the key subgraphs $\hat{G}_s \subseteq G_s$ and $\hat{G}_t \subseteq G_t$, and their contributing features $\hat{X}_s \subseteq X_s$ and $\hat{X}_t \subseteq X_t$. These subgraphs and features must enable the NA model to retain its prediction $(u \equiv v)$ when trained on the reduced subgraphs $\hat{G}_s$, $\hat{G}_t$ with the selected node features $\hat{X}_s$, and $\hat{X}_t$.

\textsf{NAEx} draws inspiration from GNN explainer models \cite{YingBYZL19}, which explain GNN predictions \( Y \) by identifying a subgraph \( \hat{G} \subseteq G \) and/or a subset of features \( \hat{X} \subseteq X \) that contribute to the predictions. While the original model predicts \( Y \) using the full graph \( G \) and all attributes \( X \), the explainer predicts \( \hat{Y} \) using \( \hat{G} \) and \( \hat{X} \).

To ensure that the extracted subgraph and features effectively capture the critical elements influencing the GNN's predictions, a common objective function~\cite{YingBYZL19} is based on the mutual information between the predictions $Y$ and $\hat{Y}$.  Formally, treating $Y$ and $\hat{Y}$ as random variables, the objective function is defined as:
\begin{eqnarray}
    \operatorname*{argmin}_{\hat{G}, \hat{X}}  - I\left(Y(G,X); \hat{Y}(\hat{G},\hat{X})\right), 
\end{eqnarray}
where the mutual information $I(Y;\hat{Y})$ is given as 
\begin{equation}
I(Y;\hat{Y}) = \sum_{y} \sum_{\hat{y}} P(Y=y, \hat{Y}=\hat{y}) 
\log \frac{P(Y=y, \hat{Y}=\hat{y})}{P(Y=y)P(\hat{Y}=\hat{y})}
\end{equation}


This term measures how well $\hat{G}$ captures the information relevant to the original prediction $Y$.

However, the above formulation is not directly suitable for the challenges of NA explanations outlined in Section \ref{sec:intro}, primarily due to the the lack of architectural capacity to model the cross-network dependencies that are central to the NA task. 

\begin{figure}[t]
\centering
\includegraphics[width=\columnwidth]{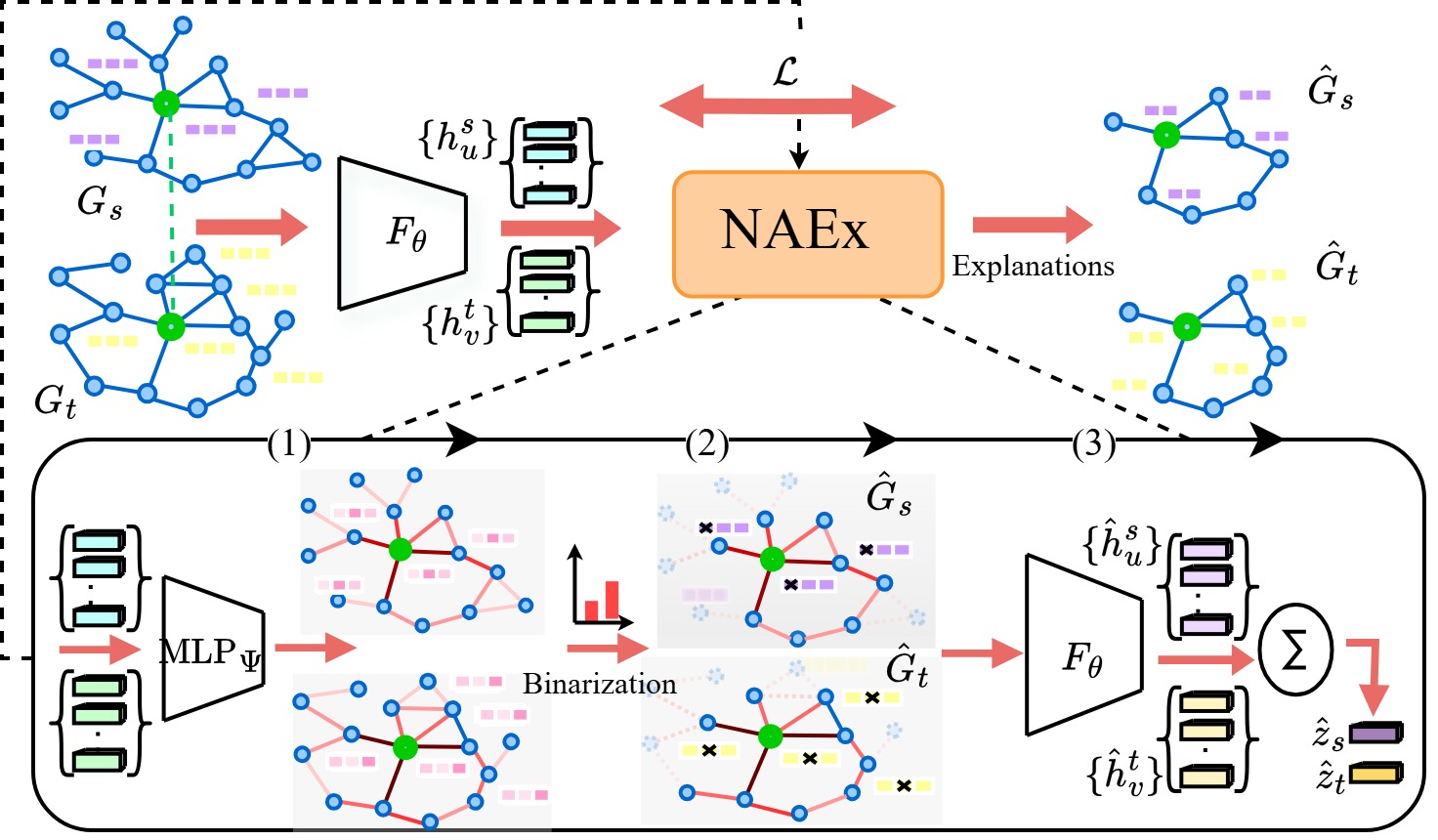}
    
\caption{ The \textsf{NAEx} framework. The upper part demonstrates the integration of \textsf{NAEx} with an NA model. (1) The first stage of edge and feature importance modeling (Sec. \ref{sec:para}). It takes nodes representation $\{{h}^s_u\}$,  $\{{h}^s_v\}$ and the original graphs ($G_s$, $G_t$) as inputs to an MLP to compute edge relevance scores and represent graphs as edge distributions. It also learns feature masks. (2) The second stage of subgraph sampling (Sec. \ref{sec:samp}). Top-ranked edges and features are selected according to edge relevance scores and feature importance. Random graphs $\hat{G}_s$ and $\hat{G}_t$ are sampled from the edge distributions and fed to the trained $F_{\theta}$ to get the node embeddings $\{\hat{h}^s_u\}$,  $\{\hat{h}^s_v\}$ and their corresponding graph embeddings ($\hat{z}_S$, $\hat{z}_t$) via a pooling function. (3) \textsf{NAEx} is optimized with the loss $\mathcal{L}$ between the original and the updated embeddings (Sec \ref{sec:opt}). }
\label{fig:NAEx_archi}
\vspace{14pt}%
\end{figure}

\section{The \textsf{NAEx} Framework}

We introduce \textsf{NAEx}, a framework for explaining network alignment (see Figure \ref{fig:NAEx_archi}). 
First, we formulate the explanation objective. 
Next, we detail the construction of explanatory subgraphs using learnable masks that capture the key substructures and features in the source and target graphs driving the alignment. Finally, we describe the optimization framework
to refine these subgraphs effectively. 


\subsection{Network Alignment Explanation Objective}
Let \( Y \) represent the prediction for a node pair \( (u, v) \) based on source and target graphs \( G_s, G_t \) and their features \( X_s, X_t \). Here, \( Y = 1 \) if \( u \equiv v \) (i.e., \( u \) aligns with \( v \)), and \( Y = 0 \) otherwise. An NA explainer predicts \( \hat{Y} \) for \( (u, v) \) using reduced subgraphs \( \hat{G}_s, \hat{G}_t \) and reduced features \( \hat{X}_s, \hat{X}_t \). The mutual information-based objective for NA can then be defined as:
\begin{eqnarray}
\begin{aligned}
\operatorname*{argmin}_{\hat{G}_s, \hat{G}_t, \hat{X}_s, \hat{X}_t} -I(Y; \hat{Y})
 = \operatorname*{argmin}_{\hat{G}_s, \hat{G}_t, \hat{X}_s, \hat{X}_t} -[H(Y) - H(Y \mid \hat{Y})]\nonumber
\end{aligned}
\end{eqnarray}

Since $H(Y)$ is fixed for a trained model, the above objective is equivalent to the optimization of 
\begin{equation}
\operatorname*{argmin}_{\hat{G}_s,\hat{G}_t,\hat{X}_s,\hat{X}_t} H(Y|\hat{Y}),
\label{eq:obj}
\end{equation}
which is to minimize the conditional entropy such that the learned explanatory subgraphs maximally preserve the alignment prediction. 

\subsection{Edge and Feature Importance Modeling}\label{sec:para}
Optimizing Equation~\ref{eq:obj} requires calculating the probabilities $P(\hat{Y})$ over all possible subgraph instances $\hat{G}_s$ and $\hat{G}_t$.  However, directly optimizing it is computationally infeasible due to the exponential growth in subgraph combinations (\( 2^{|E_s| + |E_t|} \)).
To overcome this, we adopt a sampling-based optimization approach similar to \cite{luo2020pgexplainer}, which assumes that the graph edges are conditionally independent of each other. In this work the authors parameterize a probabilistic distribution of the edges to sample subgraphs and optimize the conditional entropy based on the expected sampled graphs. 
However, our task requires simultaneous subgraph sampling from both networks to preserve cross-network dependencies. To handle this, we jointly parameterize the graph structures and feature spaces using learnable edge and feature masks. The masks identify and remove edges and features with minimal impact on predictions, allowing for efficient sampling of relevant subgraphs and features while maintaining computational feasibility.



%
\subsubsection{Graph parametrization}
We represent the graph structure as a probabilistic distribution over edges to measure the contribution of each edge to the model's predictions. Specifically, for each edge $(i, j)$ in the graphs $G_s$ and $G_t$, we learn edge masks represented by matrices $Z^{(s)} \in \{0,1\}^{|V_s| \times |V_s|}$ and $Z^{(t)} \in \{0,1\}^{|V_t| \times |V_t|}$, respectively. Thus the probabilities of graphs $G_s$ and $G_t$ based on the edge mask probabilities can be represented as
\begin{align}
P(G_s) = \prod_{(i, j) \in E_s} P(Z_{ij}^{(s)})\\
P(G_t) = \prod_{(i, j) \in E_t} P(Z_{ij}^{(t)})
\end{align}

These edge masks are specific to the given node pair $(u, v)$ for which the model's alignment prediction needs to be explained. If the edge mask is active ($Z_{ij}=1$), then the edge is selected, else not.

\subsubsection{Modeling Edge Importance}
We compute the probability of an edge mask being active, \( P(Z_{ij} = 1) \), based on an edge importance score \( \alpha_{ij} \). This score \( \alpha_{ij} \) is calculated for each edge in $E_{s}$ and $E_{t}$. For an edge $(i,j) \in E_s$, $\alpha_{ij}$ is based on the importance of $(i,j)$ w.r.t node $u$ and is calculated using the node representations $h^s_i$, $h^s_j$, and $h^s_u$. Similarly, for an edge $(i,j) \in E_t$, $\alpha_{ij}$ is calculated using the node representations $h^t_i$, $h^t_j$, and $h^t_v$. Mathematically, the importance score of an edge $(i,j) \in E_s$ w.r.t $u$ is given as:




\begin{equation}
    \alpha_{ij} = \operatorname*{MLP}_{\Psi}([h_i|| h_j|| h_{u} ])
    \label{eq:impsc}
\end{equation}
where $||$ is the concatenation operation. We share the parameters of the same MLP across $G_s$ and $G_t$ to equip \textsf{NAEx} with a global view for generating explanations. By including the representation of the source or target node as context, we ensure that edge importance is not solely determined by the local node pair $(i,j)$, but also considers how relevant the edge is in context of the alignment between $(u,v)$.

To generate binary mask values $Z_{ij}$ from the parameters $\alpha_{ij}$, we relax the binary values to continuous ones and apply a reparameterization trick to enable gradient-based backpropagation. Building on the approach in~\cite{luo2020pgexplainer}, we utilize the binary concrete distribution, parameterized by $\alpha_{ij}$, to produce $Z_{ij}$, as follows:
\begin{align}
   Z_{ij} = \sigma\Big((\log(\epsilon / (1 - \epsilon)) + \alpha_{ij}) / \beta\Big),
    \label{eq:mask}
\end{align}
where $\epsilon \sim \text{Uniform}(0, 1)$ is the random variable for sampling and $\beta$ is a temperature hyperparameter. A low $\beta$ produces a harder mask, making $Z_{ij}$ more binary-like, while a high $\beta$ yields smoother, more continuous values in $[0,1]$. 


\subsubsection{Modeling Feature Selection}
We learn a shared feature mask vector $F \in \{0,1\}^{d_m}$, which is applied to $X$ to generate $\hat{X}$ as $\hat{X} = X \odot F$, where $\odot$ denotes the Hadamard product. To enable gradient backpropagation through the binary masks, we adopt the reparameterization trick proposed in~\cite{YingBYZL19}. Specifically, we reparameterize $X$ as:  
\begin{eqnarray}
\hat{X} = W + (X - W) \odot F,
\label{eq:fmask}
\end{eqnarray}
where $W$ is a $d_m$ dimensional random variable sampled from a Gaussian distribution, parameterized by its mean and standard deviation, given as $\Omega=(\mu,\sigma^2)$. Additionally, $K_F$ is a hyperparameter that constrains the maximum number of features retained in the explanation, ensuring $\sum_j F_j \leq K_F$. This approach facilitates the learning of meaningful feature masks while maintaining computational efficiency and interpretability.

\subsection{Subgraph sampling and Explanation Generation}\label{sec:samp}
With the above graph parameterization and feature masking, the objective in Equation \ref{eq:obj} can be expressed as:
\begin{equation}
\begin{aligned}
\operatorname*{argmin}_{\hat{G}_s, \hat{G}_t, \hat{X}_s, \hat{X}_t} 
& \, H(Y|\hat{Y})
\approx 
& \operatorname*{min}_{\Psi,\Omega} 
\mathbb{E}_{\epsilon} \mathbb{E}_{W}
H(Y|\hat{Y})
\end{aligned}
\end{equation}

By sampling $\epsilon$ and $W$, the edge masks $(Z^{(s)}$ and $Z^{(t)})$ can be generated by Equation \ref{eq:mask} and the feature masks ($F_s$ and $F_t$) from Equation \ref{eq:fmask}, respectively. For efficient optimization, we modify the conditional entropy with cross-entropy $H(Y,\hat{Y})$ \cite{YingBYZL19} and adopt Monte Carlo sampling to approximate the expectation as:

\begin{align}
\operatorname*{min}_{\Psi,\Omega} -\frac{1}{K} \sum_{k=1}^{K} \Big[ 
& P(Y = 1 \mid G_s, G_t) 
\log P(\hat{Y} = 1 \mid \hat{G}_s^{(k)}, \hat{G}_t^{(k)}) \notag \\
& + P(Y = 0 \mid G_s, G_t) 
\log P(\hat{Y} = 0 \mid \hat{G}_s^{(k)}, \hat{G}_t^{(k)}) 
\Big]
\label{eq:cross}
\end{align}
where $\hat{G}_s$, $\hat{G}_t$ consist of the most relevant edges selected by applying the obtained masks $Z^{(s)}$ and $Z^{(t)}$,
\begin{equation}
    \hat{E}_s= E_s \odot Z^{(s)}, \quad \hat{E}_t= E_t \odot Z^{(t)}  
\end{equation}
along with their non-masked features $\hat{X}_s$ and $\hat{X}_t$.
\subsection{Optimization Framework}\label{sec:opt}
The optimization of the above objective in Equation \ref{eq:cross} ensures that the explanations are faithful to the alignment model. To further ensure that the explanations are not random but grounded in meaningful subgraph structures, we introduce a subgraph similarity objective. This encourages consistency in the explanations for aligned nodes, improving their interpretability. Additionally, we incorporate regularization terms to penalize overly complex or dense explanations Then, we have the general loss function of \textsf{NAEx} for generating the explanations of an anchor $(u,v)$ as:
\begin{equation}
\mathcal{L}^{\text{Ex}}=-\frac{1}{K} \sum_{k=1}^{K} \Big[ \mathcal{L}^{align}+ \lambda_1 \mathcal{L}^{subgraph}+\lambda_2 \left (\mathcal{L}^{\text{feat}}+\mathcal{L}^{\text{edge}}  \right)\Big]
\end{equation}
where $\mathcal{L}^{align}$ is the alignment consistency loss that approximates the cross-entropy objective, $\mathcal{L}^{subgraph}$ is the subgraph similarity loss that enforces similarity in the explanation subgraphs, and $\mathcal{L}^{\text{feat}}$, $\mathcal{L}^{\text{edge}}$ are feature and edge sparsity regularizers respectively.
To provide collective explanations for multiple instances $(u,v) \in {\mathcal{I}}$, where $I \subseteq A$, the objective of \textsf{NAEx} becomes:
\begin{equation}
    \mathcal{L} = \sum_{(u,v) \in \mathcal{I}}  \mathcal{L}^{\text{Ex}}
    \label{eq:FLoss}
\end{equation}

We suggest readers to refer to Algorithm 1 in the Supplementary, which summarizes the training process of \textsf{NAEx}.

\noindent \underline{\textbf{Computational Complexity.}} \textsf{NAEx} is efficient for two reasons. First, the explanation network $\Psi$ in \textsf{NAEx}, which learns a latent variable for each edge, is shared across all the edges of $G_s$ and $G_t$. Thus, a trained \textsf{NAEx} can explain new instances in the inductive setting in $O(|E_s|+|E_t|)$, making it scalable to large graphs. Second, since the features are shared among the nodes across the network, the learned feature explanation is naturally global and applicable to new instances in the inductive setting. During training, \textsf{NAEx} maintains comparable complexity to existing GNN-based explanation methods (\cite{YingBYZL19,luo2020pgexplainer}) by efficiently reusing parameters across edges and features, despite operating on network pairs. Moreover, empirical results demonstrate that \textsf{NAEx} requires training on only a small subset of the data, further improving training efficiency (refer Section \ref{sec:quant}).

\subsubsection{Alignment consistency loss}
We reformulate the cross entropy objective of Equation \ref{eq:cross} with the help of alignment probabilities. 
We leverage the similarity between node embeddings to approximate the alignment predictions. For a given node pair $(u,v)$, we have 
\begin{align}
P(Y = 1 \mid G_s, G_t) &\approx \sigma\left(\text{sim}(h_u^s, h_v^t)\right) = s_{uv} \\
P(\hat{Y} = 1 \mid \hat{G}_s, \hat{G}_t) &\approx \sigma\left(\text{sim}(\hat{h}_u^s, \hat{h}_v^t)\right) = \hat{s}_{uv}
\label{eq:pred}
\end{align}
where $\hat{h}^s_u=F_{\theta}(u | \hat{G_s})$ and $\hat{h}^t_v=F_{\theta}(v | \hat{G_t})$. $s_{uv}$ is the true alignment score for $(u,v)$ and $\hat{s}_{uv}$ is the predicted alignment score for $(u,v)$ given $\hat{G}_s$ and $\hat{G}_t$. The sigmoid function $\sigma$ transforms the similarity into a probabilistic score, ensuring that the result is a valid probability within the range $[0,1]$.
Integrating negative sampling into the objective to improve training stability and efficiency, gives:
\begin{equation}
\begin{aligned}
L^{\text{align}}_{(u, v)} = 
&\ s_{uv} \log \hat{s}_{uv} + \sum_{(u', v') \not\in A} s_{u'v'} \log \hat{s}_{u'v'}.
\end{aligned}
\end{equation}

This formulation ensures that the explanatory subgraphs are optimized to reflect the alignment process for the positive pair, while reducing alignment for negative samples. From a global perspective, it also brings the notion of the degree of alignment or uncertainty. Instead of over-fitting to uncertain positives or hard negatives, the model learns to focus on the most reliable signals when tested for a new node pair. Hence, the inductive as well as the global explanation capability of the model is enhanced.

\subsubsection{Subgraph similarity loss}

While $L^{align}$ focuses on edges and features that directly contribute to improving node alignment probabilities at the pair-wise level, we also introduce a subgraph-level contrastive loss. It optimizes the selection of edges and features that make the explanatory subgraphs similar for the aligned node pairs.
\begin{eqnarray}
L_{(u,v)}^{\text{contrastive}} = 
\log \frac{e^{\text{sim}(\hat{z}_s, \hat{z}_t) / \tau}}
{e^{\text{sim}(\hat{z}_s, \hat{z}_t) / \tau}
+ \sum_{G^{\star} \neq \hat{G}_t} e^{\text{sim}(\hat{z}_s, z^{\star}_t) / \tau}}.
\end{eqnarray}

The graph embeddings $\hat{z}_s$ and $\hat{z}_t$ represent the pooled node embeddings of the explanatory subgraphs $\hat{G}_s$ and $\hat{G}_t$, which are sampled with respect to the nodes $u$ and $v$, respectively. These embeddings are defined as $\hat{z}_s = \tanh\left(\frac{1}{n} \sum_{i=1}^n \hat{h}^s_i\right)$. The contrastive loss encourages embeddings of explanatory subgraphs for positive pairs $(u,v)$ to be closer, while pushing embeddings for negative pairs farther apart. To make the subgraph embeddings similar for the aligned pairs, it must focus on selecting edges that highlight alignment- specific edges. Similarly, we consider the $L_{(v,u)}^{\text{contrastive}}$ to derive the overall function of the subgraph level contrastive loss as:
\begin{eqnarray}
\label{eq:globalcon}
\mathcal{L}^{\text subgraph}=\frac{1}{2}\left[L^{contrastive}_{(u,v)}+L^{contrastive}_{(v,u)}\right]
\end{eqnarray}
\subsubsection{Regularization}
To ensure compact and interpretable explanations, we impose sparsity constraints on both feature masks and edge importance scores:

\noindent\textbf{\textit{Feature sparsity:}} We regularize the  $L_1$-norm of the feature mask to encourage the model to focus on a smaller subset of relevent features:
\begin{equation}
    \mathcal{L}^{\text{feat}} = \|F \|_1 
\end{equation}


\noindent\textbf{\textit{Edge sparsity:}} We impose an \( L_1 \)-norm penalty on the edge importance scores \( \alpha_{ij} \) for both the source and target graphs to encourage the selection of a sparse subset of edges:
\begin{equation}
    \mathcal{L}^{\text{edge}} =\sum_{(i,j) \in E^s} |\alpha_{ij}^{(s)}| + \sum_{(i,j) \in E^t} |\alpha_{ij}^{(t)}|
\end{equation}

\begin{table}[t]
\caption{Dataset statistics}
\centering
\setlength{\extrarowheight}{4pt}
\label{tab:my-table}
\resizebox{\columnwidth}{!}{%
\begin{tabular}{l|c|c|c|c}
\hline
Dataset          & $|$V$|$           & $|$E$|$             & \#Features & \#Anchors \\ \hline
Foresq - Twitter (F-T)& 5,313 - 5,120 & 76,972 - 1,64,919 & 0           & 1,609          \\
ACM - DBLP (A-D)      & 9,916 - 9,872   & 44,808 - 39,561   & 17          &   6,325       \\
Allmovie - IMDB (A-I) & 6,011-5,713     & 1,24,709 - 1,19,073 & 14          & 5,176     \\ \hline
\end{tabular}%
}
\label{tab:datasets}
\end{table}

\section{Experimental Setup}


\subsection{Datasets}

We evaluate the performance of \textsf{NAEx} using three benchmark real-world datasets from diverse domains. These include social networks (Foursquare-Twitter) \cite{seo2024leveraging}, movie guide service networks (AllMovie-IMDB) \cite{trung2020adaptive}, and co-authorship networks (ACM-DBLP) \cite{seo2024leveraging}, where nodes represent users, films, and authors, respectively. The dataset statistics are provided in Table \ref{tab:datasets}. For Foresquare-Twitter dataset, we use Laplacian eigenvectors \cite{dwivedi2023benchmarking} as node features.

\subsection{Baselines}
To comprehensively evaluate the effectiveness of \textsf{NAEx}, we compare it with several post hoc explanation strategies:
\begin{enumerate}[leftmargin=*]
    \item \textbf{Random} \cite{huang2022graphlime} assigns random importance scores to edges and features and serves as a naive baseline for comparison.
    \item \textbf{Grad} \cite{YingBYZL19} computes the gradient of the NA model's loss function with respect to the adjacency matrix and the associated node features, similar to the saliency map method.
    \item \textbf{dNAME} \cite{zhou2019disentangled}: uses influence functions to estimate the importance of each node for aligning a given anchor pair $(u, v)$. It uses a linear approximation to estimate the parameter changes when a data point is removed.
    \item \textbf{Atten} \cite{velickovic2017graph}  leverages the attention weights from GAT based encoders to identify important edges. It does not consider node features and is limited to NA mdoels using attention mechansims.
    \item \textbf{GNNExplainer} \cite{YingBYZL19} is a GNN explanation model that identifies a compact subgraph and feature subset maximizing the mutual information with the model's prediction. We adapt it for NA by jointly learning node and feature masks over the local neighborhoods of the aligned pair in both graphs. 
\end{enumerate}

\subsection{Applying \textsf{NAEx} to NA methods}
\textsf{NAEx} can be directly applied to any existing deep neural embedding-based NA model. The framework leverages the parameters of the trained encoder used for generating node embeddings, enabling seamless integration with the model. Some state-of-the-art NA models that can be explained by \textsf{NAEx} include - GAlign \cite{trung2020adaptive}, LSNA \cite{liang2021unsupervised}, JORA \cite{zheng2022jora}, Grad-Align+ \cite{park2022gradalign+}, HTC \cite{sun2023towards}, NAME \cite{huynh2021network}, MINING \cite{zhang2023mining}, SAlign \cite{saxena2023salign}, ASSISTANT \cite{seo2024leveraging}, SANA \cite{peng2023robust}, NeXtAlign \cite{zhang2021balancing}, AMN \cite{li2025attention} and many more.

Due to the limitation of space, we only show experimental results for four NA models namely JORA \cite{zheng2022jora}, SANA \cite{peng2023robust}, HTC \cite{sun2023towards}, and ASSISTANT \cite{seo2024leveraging} in conjunction with our work. More details about these methods can be found in Supplementary. 

\begin{table}[t]
\caption{Performance of \textsf{NAEx} in transductive setting}
\centering
\setlength{\extrarowheight}{2.2pt}
\resizebox{0.7\columnwidth}{!}{%
\begin{tabular}{c|c|c|c|c}
\hline
\multirow{2}{*}{Dataset} & 
\multirow{2}{*}{NA model} & 
\multirow{2}{*}{\begin{tabular}[c]{@{}c@{}}FID \end{tabular}} & 
\multirow{2}{*}{\begin{tabular}[c]{@{}c@{}}FTH \end{tabular}} & 
\multirow{2}{*}{\begin{tabular}[c]{@{}c@{}}Spar \end{tabular}} \\ 
                          &           &       &       &       \\ \hline
\multirow{4}{*}{Foresq - Twitter} & JORA      & 0.711 & 0.571 & 0.513 \\ 
                                  & SANA      & 0.728 & 0.585 & 0.504 \\ 
                                  & HTC       & \textbf{0.788} & 0.601 & \textbf{0.538} \\ 
                                  & ASSISTANT & 0.783 & \textbf{0.668} & 0.512 \\ \hline
\multirow{4}{*}{ACM-DBLP}         & JORA      & 0.773 & 0.751 & 0.562 \\ 
                                  & SANA      & 0.782 & 0.738 & 0.505 \\ 
                                  & HTC       & 0.811 & \textbf{0.782} & \textbf{0.595} \\ 
                                  & ASSISTANT & \textbf{0.824} & 0.774 & 0.582 \\ \hline
\multirow{4}{*}{Allmovie-IMDB}    & JORA      & 0.882 & 0.794 & 0.665  \\ 
                                  & SANA      & 0.871 & 0.806 & 0.662 \\ 
                                  & HTC       & 0.885 & \textbf{0.828} & \textbf{0.695} \\ 
                                  & ASSISTANT & \textbf{0.893} & 0.822 & 0.682 \\ \hline
\end{tabular}%
}
\label{tab:trans}
\vspace{9pt}%
\end{table}

\begin{table}[t]
\caption{Baseline comparison. S denotes the sparsity}
\label{tab:baselines}
\setlength{\extrarowheight}{2.2pt}
\resizebox{\columnwidth}{!}{%
\begin{tabular}{c|c|ccc|ccc|ccc}
\hline
\multirow{2}{*}{Baselines} &
  \multirow{2}{*}{Metric} &
  \multicolumn{3}{c|}{Foresq-Twitter} &
  \multicolumn{3}{c|}{ACM-DBLP} &
  \multicolumn{3}{c}{AllMovie-IMDB} \\ \cline{3-11} 
 &
   &
  \multicolumn{1}{c|}{S=0.5} &
  \multicolumn{1}{c|}{S=0.6} &
  S=0.7 &
  \multicolumn{1}{c|}{S=0.5} &
  \multicolumn{1}{c|}{S=0.6} &
  S=0.7 &
  \multicolumn{1}{c|}{S=0.5} &
  \multicolumn{1}{c|}{S=0.6} &
  S=0.7 \\ \hline
\multirow{2}{*}{\textbf{Random}} &
  FID &
  \multicolumn{1}{c|}{0.270} &
  \multicolumn{1}{c|}{0.227} &
  0.194 &
  \multicolumn{1}{c|}{0.298} &
  \multicolumn{1}{c|}{0.265} &
  0.241 &
  \multicolumn{1}{c|}{0.375} &
  \multicolumn{1}{c|}{0.356} &
  0.311 \\ 
 &
  FTH &
  \multicolumn{1}{c|}{0.218} &
  \multicolumn{1}{c|}{0.191} &
  0.163 &
  \multicolumn{1}{c|}{0.230} &
  \multicolumn{1}{c|}{0.204} &
  0.199 &
  \multicolumn{1}{c|}{0.328} &
  \multicolumn{1}{c|}{0.287} &
  0.275 \\ \hline
\multirow{2}{*}{\textbf{Grad}} &
  FID &
  \multicolumn{1}{c|}{0.658} &
  \multicolumn{1}{c|}{0.633} &
  0.528 &
  \multicolumn{1}{c|}{0.723} &
  \multicolumn{1}{c|}{0.689} &
  0.599 &
  \multicolumn{1}{c|}{0.840} &
  \multicolumn{1}{c|}{0.806} &
  0.737 \\
 &
  FTH &
  \multicolumn{1}{c|}{0.517} &
  \multicolumn{1}{c|}{0.441} &
  0.417 &
  \multicolumn{1}{c|}{0.635} &
  \multicolumn{1}{c|}{0.564} &
  0.511 &
  \multicolumn{1}{c|}{0.718} &
  \multicolumn{1}{c|}{0.652} &
  0.594 \\ \hline
\multirow{2}{*}{\textbf{dNAME}} &
  FID &
  \multicolumn{1}{c|}{0.672} &
  \multicolumn{1}{c|}{0.635} &
  0.518 &
  \multicolumn{1}{c|}{0.757} &
  \multicolumn{1}{c|}{0.724} &
  0.628 &
  \multicolumn{1}{c|}{0.843} &
  \multicolumn{1}{c|}{0.827} &
  0.778 \\
 &
  FTH &
  \multicolumn{1}{c|}{0.522} &
  \multicolumn{1}{c|}{0.476} &
  0.429 &
  \multicolumn{1}{c|}{0.638} &
  \multicolumn{1}{c|}{0.581} &
  0.522 &
  \multicolumn{1}{c|}{0.739} &
  \multicolumn{1}{c|}{0.693} &
  0.640 \\ \hline
\multirow{2}{*}{\textbf{Atten}} &
  FID &
  \multicolumn{1}{c|}{0.542} &
  \multicolumn{1}{c|}{0.441} &
  0.408 &
  \multicolumn{1}{c|}{0.646} &
  \multicolumn{1}{c|}{0.608} &
  0.543 &
  \multicolumn{1}{c|}{0.694} &
  \multicolumn{1}{c|}{0.622} &
  0.549 \\
 &
  FTH &
  \multicolumn{1}{c|}{0.500} &
  \multicolumn{1}{c|}{0.412} &
  0.396 &
  \multicolumn{1}{c|}{0.601} &
  \multicolumn{1}{c|}{0.547} &
  0.509 &
  \multicolumn{1}{c|}{0.677} &
  \multicolumn{1}{c|}{0.583} &
  0.514 \\ \hline
\multirow{2}{*}{\textbf{GNNExp}} &
  FID &
  \multicolumn{1}{c|}{0.692} &
  \multicolumn{1}{c|}{0.654} &
  0.573 &
  \multicolumn{1}{c|}{0.771} &
  \multicolumn{1}{c|}{0.703} &
  0.649 &
  \multicolumn{1}{c|}{0.858} &
  \multicolumn{1}{c|}{0.820} &
  0.772 \\
 &
  FTH &
  \multicolumn{1}{c|}{0.528} &
  \multicolumn{1}{c|}{0.493} &
  0.451 &
  \multicolumn{1}{c|}{0.655} &
  \multicolumn{1}{c|}{0.618} &
  0.587 &
  \multicolumn{1}{c|}{0.774} &
  \multicolumn{1}{c|}{0.718} &
  0.672 \\ \hline
\multirow{2}{*}{\textbf{NAEx}} &
  FID &
  \multicolumn{1}{c|}{\textbf{0.716}} &
  \multicolumn{1}{c|}{\textbf{0.683}} &
  \textbf{0.570} &
  \multicolumn{1}{c|}{\textbf{0.785}} &
  \multicolumn{1}{c|}{\textbf{0.778}} &
  \textbf{0.738} &
  \multicolumn{1}{c|}{\textbf{0.889}} &
  \multicolumn{1}{c|}{\textbf{0.873}} &
  \textbf{0.862} \\
 &
  FTH &
  \multicolumn{1}{c|}{\textbf{0.593}} &
  \multicolumn{1}{c|}{\textbf{0.562}} &
 \textbf{ 0.527} &
  \multicolumn{1}{c|}{\textbf{0.743}} &
  \multicolumn{1}{c|}{\textbf{0.722}} &
  \textbf{0.706} &
  \multicolumn{1}{c|}{\textbf{0.816}} &
  \multicolumn{1}{c|}{\textbf{0.808}} &
  \textbf{0.804} \\ \hline
\end{tabular}%
}
\end{table}

\subsection{Evaluation metrics}
Quantitative assessment of explanation quality is challenging due to the typical absence of ground truth explanations. Therefore, we extend two widely used evaluation metrics from the literature to NA:

\noindent\textbf{Fidelity }\cite{chen2023generative}. It quantifies how well an explanation model preserves the alignment prediction of the original model when only the explanatory subgraphs are retained. 
\begin{equation}
\text{FID} = \frac{1}{|I|} \sum_{(u,v) \in I} \mathbb{I} \left[ \hat{Y}_{uv} = Y_{uv} \right]
\end{equation}
where $Y$ is the prediction given $G_S$, $G_t$ and $\hat{Y}$ is the prediction given $\hat{G}_S$, $\hat{G}_t$. $I$ represents the set of node pairs explained and $\mathbb{I}[.]$ is the indicator function. A high fidelity means the explanations correctly represent the original model's behavior.

\noindent\textbf{Faithfulness }\cite{agarwal2023evaluating}.
It refers to the extent to which the explanation represents the original model's true underlying alignment structure. It measures the difference between the prediction probability distributions of the original graphs and the explanatory subgraphs.
\begin{equation}
    \text{FTH} = \frac{1}{|I|} \sum_{(u,v) \in I} \exp^{-\text{KL}(\boldsymbol q_u |\boldsymbol{\hat{q}}_u) }
\end{equation}
where $\text{KL}(\boldsymbol q_u |\boldsymbol{\hat{q}}_u)$ is the KL diveregence between $\boldsymbol q_u=[s_{uv_1},s_{uv_2},...s_{uv_{|V_t|}}]$ and $\boldsymbol{\hat{q}}_u=[\hat{s}_{uv_1},\hat{s}_{uv_2},...\hat{s}_{uv_{|V_t|}}]$. For a candidate $v_i \not\in \hat{G}_t$, we set $\hat{s}_{uv_i}=0$ and re-normalize $\boldsymbol{\hat{q}}_u$.
Faithfulness is more granular than the fidelity metric as it quantifies subtle changes in probabilities for all alignment candidates. FTH values are expected to be comparatively lower than fidelity, as capturing the full distribution is stricter than aligning binary outcomes.

\noindent\textbf{Sparsity }\cite{chen2023generative}. It is an auxiliary metric to evaluate how compact the explanatory subgraphs ($\hat{G}_s, \hat{G}_t$) are with respect to the original networks ($ G_s, G_t$). It quantifies the relative reduction in graph size needed to explain the alignment.
\begin{equation}
    \text{Spar} = 1 -\frac{1}{|I|} \sum_{(u,v) \in I} \frac{|\hat{E}_{s_u}| + |\hat{E}_{t_v}|}{|E_{s_u}^{\star}| + |E_{t_v}^{\star}|}
\end{equation}
Here, \( \hat{E}_{s_u} \) and \( \hat{E}_{t_v} \) are the edges in the explanations for node pair \( (u, v) \), while \( E_{s_u}^\star \) and \( E_{t_v}^\star \) are the edges in their 2-hop neighborhoods. 
A higher sparsity indicates that fewer edges are retained, leading to more compact explanations.

 \begin{figure*}[!ht]
 \centering
    \begin{minipage}{0.6\linewidth}
      \subfigure[Foresq-Twitter]
        {\includegraphics[scale=0.129]{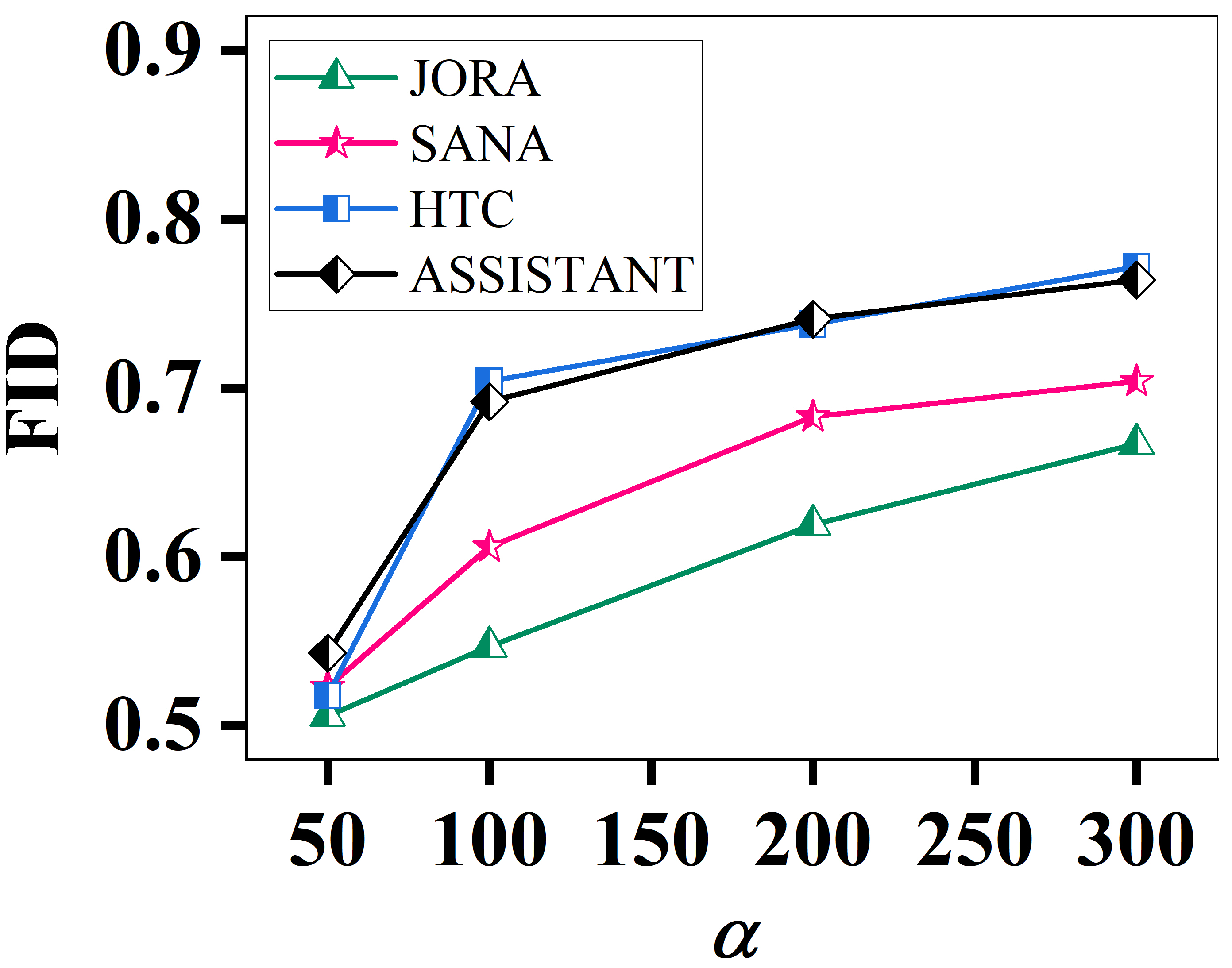}
        }
    \hspace{0.1cm}
    \subfigure[ACM-DBLP]{
        \includegraphics[scale=0.129]{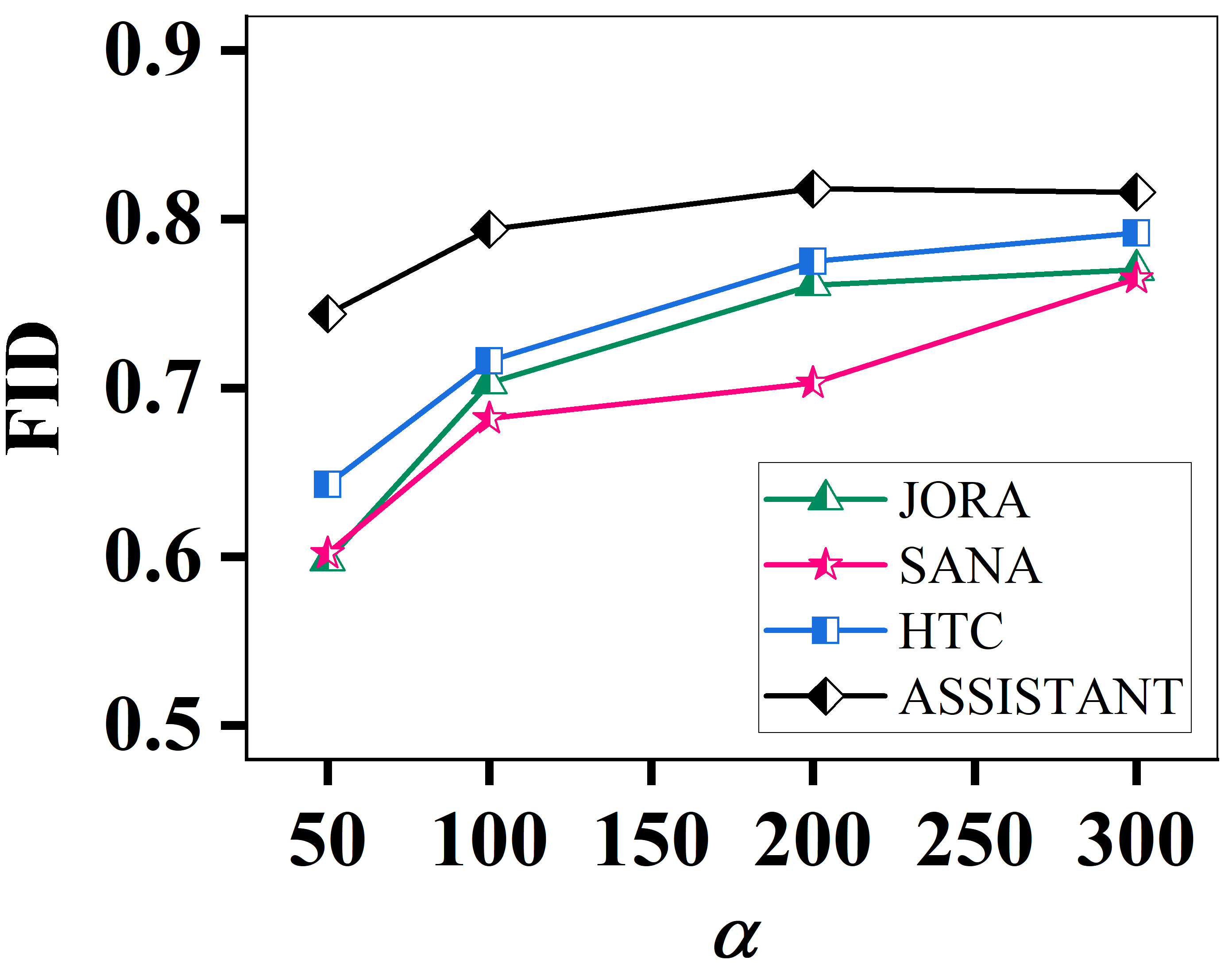}
        }
    \hspace{0.1cm}
    \subfigure[AllMovie-IMDB]
        {\includegraphics[scale=0.129]{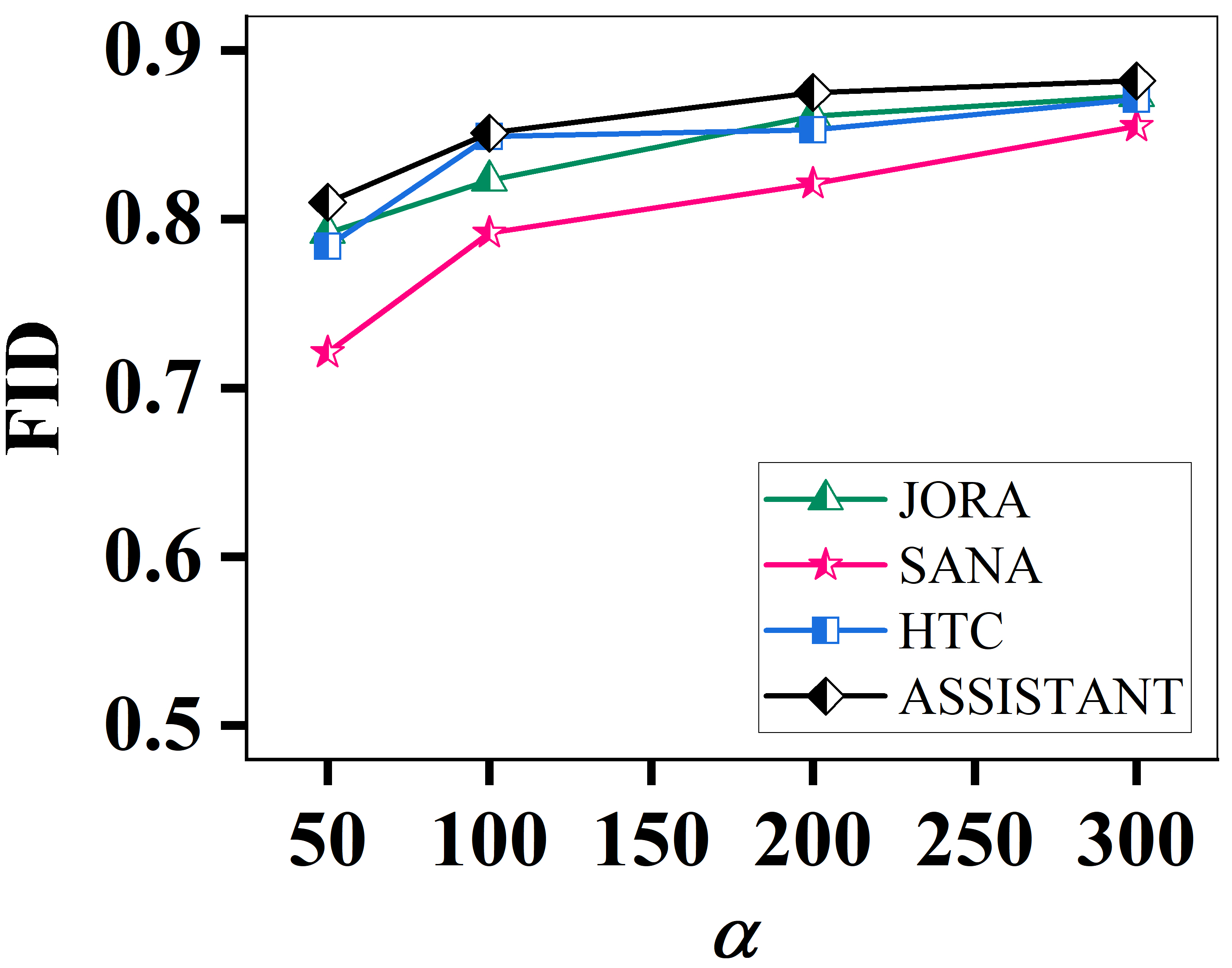}
        }
      \caption{Evalution of \textsf{NAEx} under inductive setting}\label{fig:ind}
    \end{minipage}
    \hfill
    \centering
    \begin{minipage}{0.3\linewidth}
      \includegraphics[width=45mm,height=28mm]{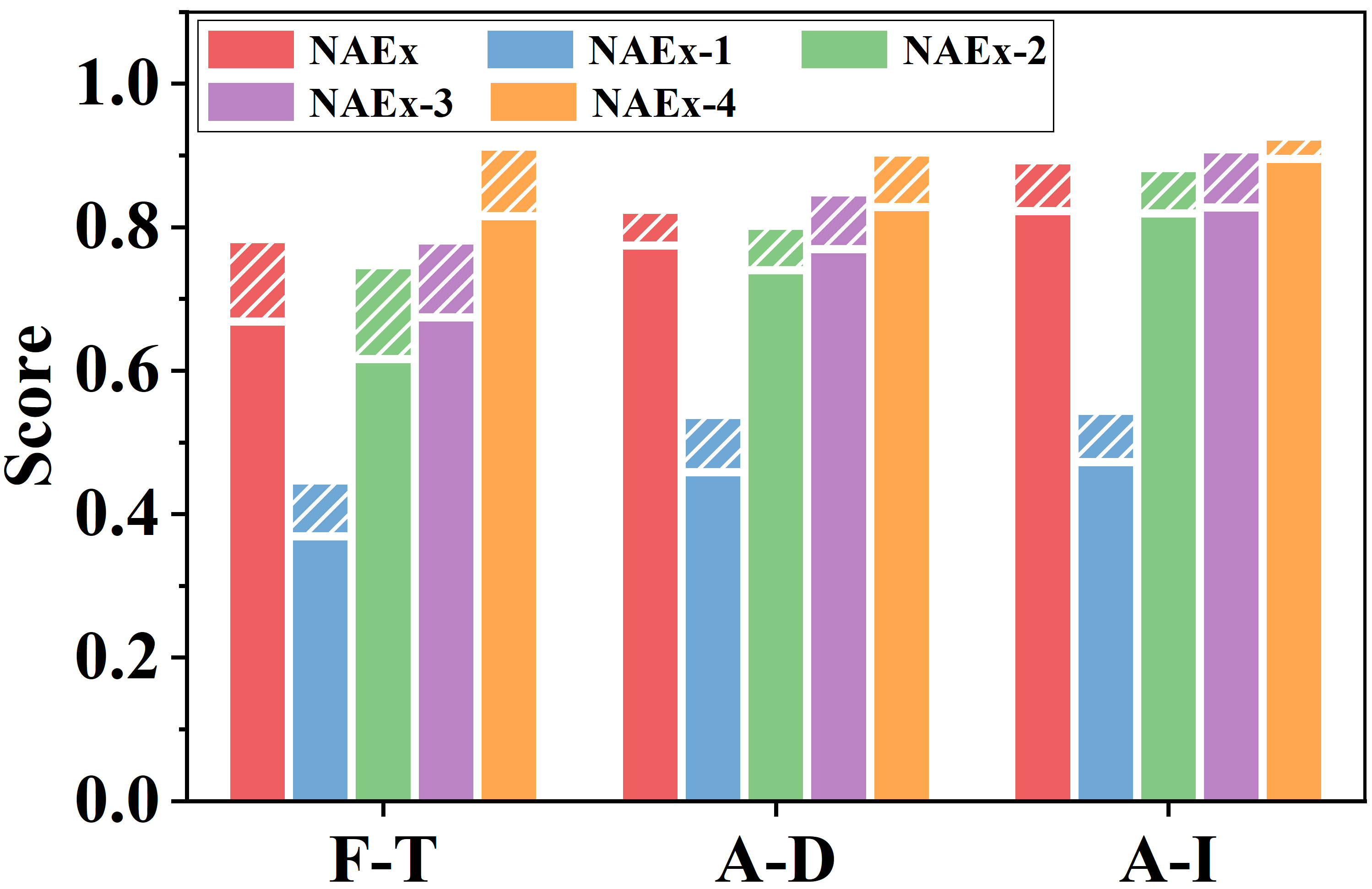}
      \caption{ Ablation study. The solid bar represents the FTH score and the dashed bar represents the corresponding FID score}\label{fig:ablation}
    \end{minipage}
  \end{figure*}

\begin{figure}[t]
\centering
    \subfigure[Training time of \textsf{NAEx}.]
        {\includegraphics[width=39mm,height=24mm]{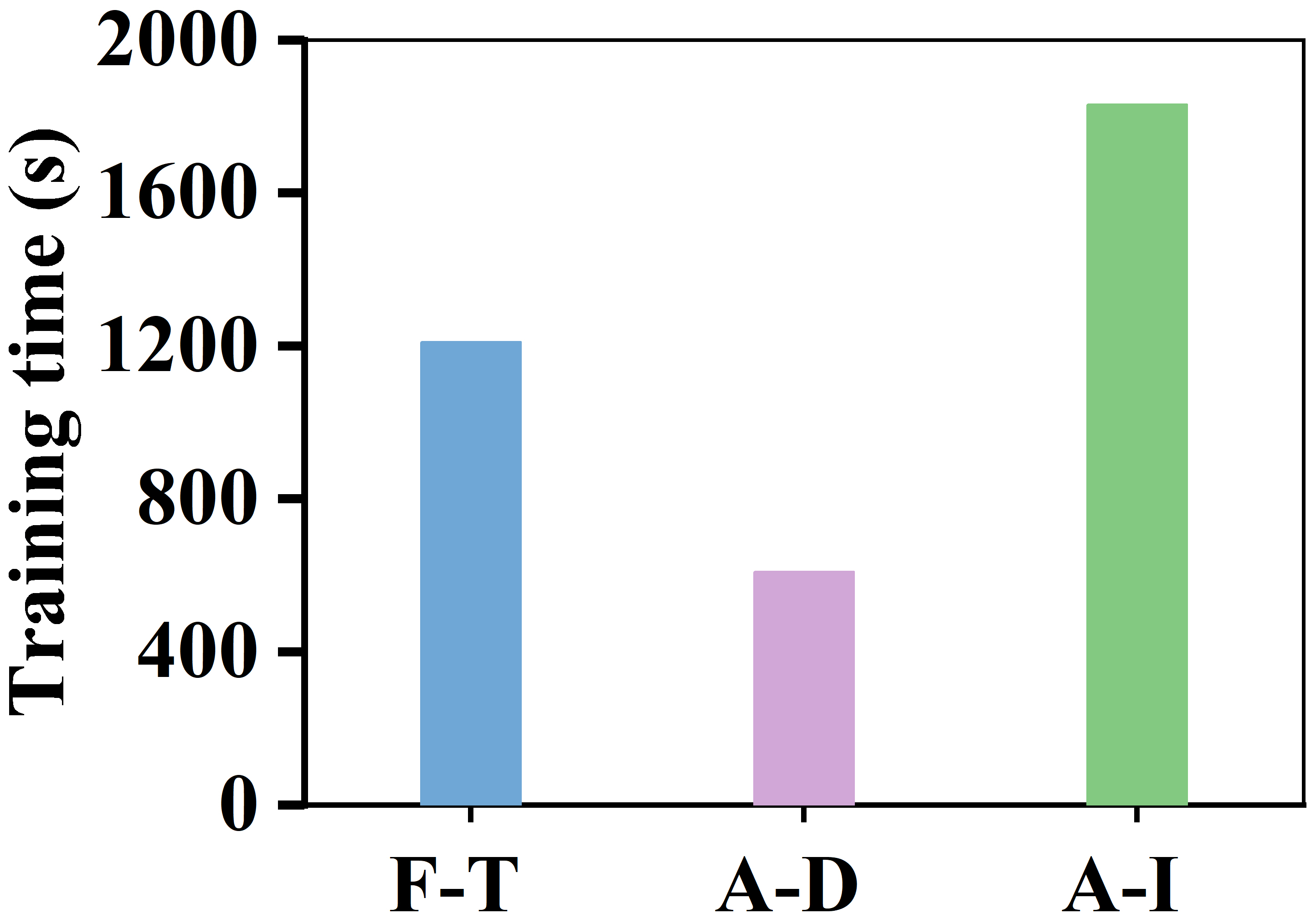}
        }
        \label{fig:training}
    \hspace{0.3cm}
    \subfigure[Inference time comparison]{
        \includegraphics[width=39mm,height=24mm]{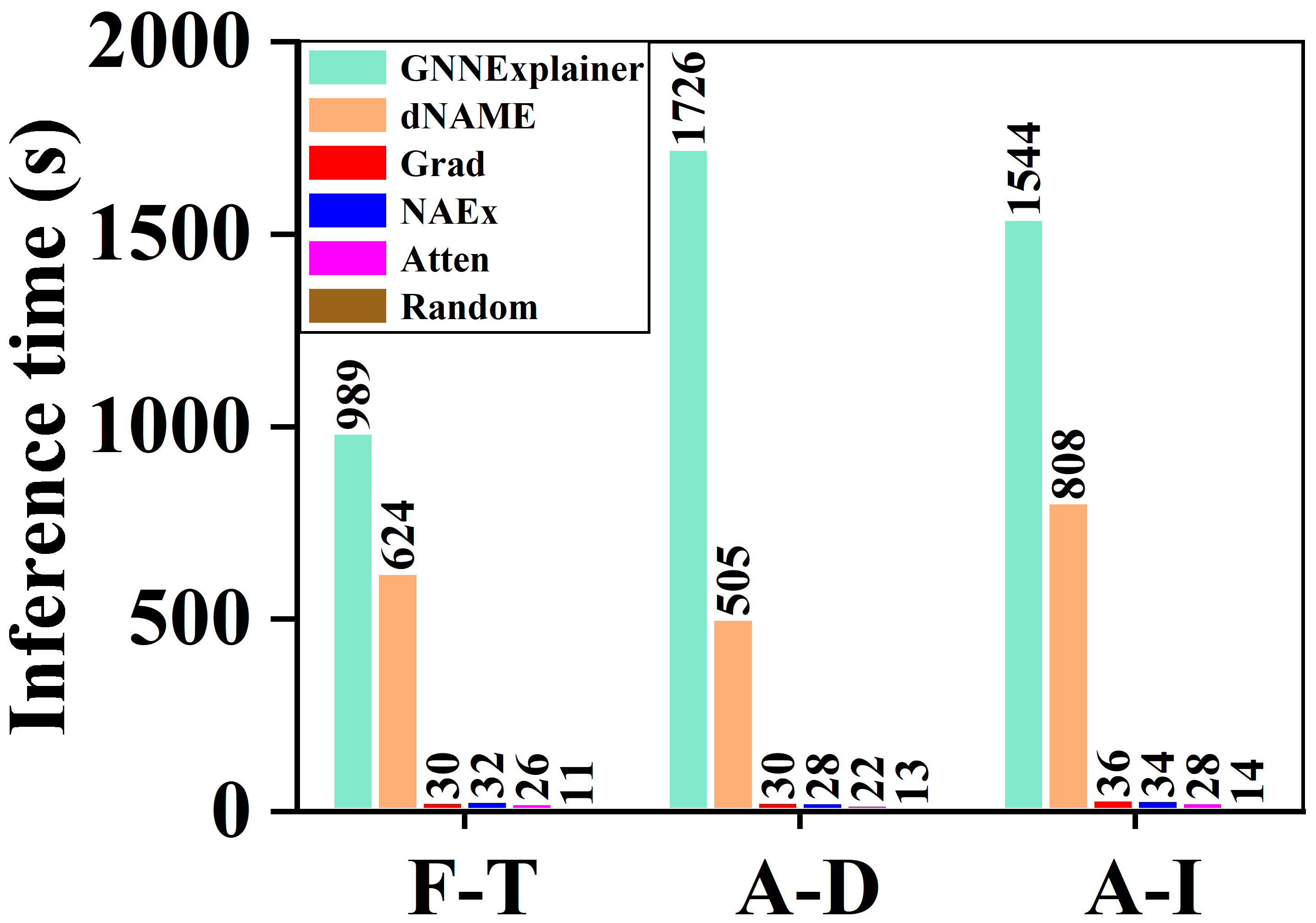}
        }
        \label{fig:infer}
\caption{Efficiency evalution}
\vspace{13pt}%
\label{fig:eff}
\end{figure}

\section{Experimental Results}
\subsection{Quantative Evaluation} \label{sec:quant}
We evaluate \textsf{NAEx} using the predictions from four NA models across three datasets. The evaluation includes two distinct settings: \textit{transductive}, where the shared explainer is trained on each anchor pair to generate explanations; and \textit{inductive}, where the pre-trained explainer is directly utilized to infer explanations for unseen anchor pairs without requiring retraining.

\noindent \underline {\textbf{Transductive setting.}} We use Submodular Pick \cite{huang2022graphlime,ribeiro2016should} to select $B$ samples whose explanations jointly maximize edge and feature coverage, enabling an efficient examination of diverse and representative model behaviors. We present the results as $B$ varies from 5 to 100, averaged over 200 rounds, in Table \ref{tab:trans}.
We see that \textsf{NAEX} performs better on the AllMovie-IMDB and ACM-DBLP datasets which have inherently rich structural and semantic overlap (e.g., shared actors or genres in AllMovie-IMDB, and shared topics between authors or papers in ACM-DBLP). These result in higher FID and FTH scores, highlighting \textsf{NAEX}’s ability to effectively capture both alignment and probability distributions. In contrast, the Foresq-Twitter dataset, which lacks attribute information and exhibits limited structural similarity shows lower FID scores and a wider gap with the FTH scores. Among the baseline NA models, ASSISTANT and HTC have superior performance as they capture the higher-order neighborhood structures in their embedding generation process. To train \textsf{NAEx} on million order datasets one can use advanced techniques for scalability in GNNs \cite{li2021training}. 

\noindent\underline {\textbf{Inductive setting.}} Since the explanation network (\(\text{MLP}_{\Psi}\)) is shared across the nodes of both networks, a trained \textsf{NAEx} model can be directly utilized to infer explanations for unseen node pairs without requiring retraining. Additionally, the selected features are consistent across the nodes of a network, making feature explanations global and applicable to new instances, even in inductive settings. For evaluating \textsf{NAEx} under different training sizes, we use \(\alpha\) anchors from the anchor set, reserve \(\frac{N - \alpha}{2}\) nodes for validation, and allocate the remaining nodes for testing. The value of \(\alpha\) is varied over \([50, 100, 150, 200, 250]\) with $N=1000$.
Results from Figure \ref{fig:ind} suggest that for all datasets, fidelity improves as the number of training instances increases. This suggests that more training data improves the ability of models to generate accurate explanations. Foresq-Twitter is the most sensitive to $\alpha$, suggesting that given its low structural consistency between the networks and no attributes, it is comparatively difficult to generalize. For other datasets, we get comparable FID at $\alpha=200$, suggesting that \textsf{NAEx} tends to detect global patterns and generalize well even with small training data.

\noindent\underline {\textbf{Baseline comparison}}. For a fair comparison, we evaluate all baseline explanation methods on top of SANA, which incorporates an attention mechanism in its encoder. We report the FID and FTH scores averaged over 50 anchor pairs across varying sparsity levels in Table~\ref{tab:baselines}. For Grad and dNAME, we calculate the importance of an edge by averaging the connected nodes' importance scores. Among the baselines, GNNExplainer and dNAME achieve moderate performance, while GRAD and Atten perform less reliably. Atten uses attention weights to reflect model focus, but these weights do not always correspond directly to high fidelity or faithful explanations. GRAD is simple and efficient, but suffers from noise, saturation, and a lack of contextual knowledge. In contrast, \textsf{NAEx} consistently outperforms all baselines, achieving an average improvement of $2.87-65.74\%$ in FID and $9.32-68.89\%$ in FTH.

\underline{\textit{Comparison with GNNExplainer}}. GNNExplainer is originally designed for node and graph classification tasks, where the model output is a discrete class label. In contrast, network alignment involves computing a continuous similarity score between all possible node pairs across two graphs. This fundamental shift in task formulation renders GNNExplainer less effective in this setting. Our empirical evaluation demonstrates that GNNExplainer exhibits notably low FID and FTH scores. More notably, its FTH scores are substanially lower, indicating that the explanations do not truly reflect the underlying reasoning of the model. Moreover, the FTH scores deteriorates rapidly as explanation sparsity increases. These shortcomings stem primarily from the fact that GNNExplainer operates on a single-graph view, lacking the architectural capacity to model cross-graph dependencies, which are central to the alignment task. In contrast, our proposed method, \textsf{NAEx}, directly optimizes explanation masks jointly over both input graphs to preserve the alignment score between node pairs. It incorporates a subgraph similarity loss to ensure that explanations are directly comparable and interpretable. This alignment-specific formulation allows \textsf{NAEx} to produce significantly more faithful and interpretable explanations that accurately reflect the matching behavior of the underlying NA model.

\subsection{Ablation study on model design}
We evaluate \textsf{NAEx} against four ablated variants: \textsf{NAEx}-1 removes alignment consistency loss, \textsf{NAEx}-2 omits subgraph similarity loss, \textsf{NAEx}-3 uses only edge masking  (while retaining all node features), and \textsf{NAEx}-4 uses only feature masking (retaining all edges). We also include a fifth variant without regularization, whose performance under varying sparsity is the same as reported in Table 4. We provide results over 50 random anchors in Figure 3. \textsf{NAEx}-1 shows the largest decline in both FID and FTH, emphasizing its importance for maintaining alignment rankings. \textsf{NAEx}-2 also underperforms, highlighting the need for structural consistency in the generated explanations. \textsf{NAEx}-3 performs similar on F-T which is a feature-sparse dataset, indicating limited feature reliance, while slightly improving on feature-rich datasets, suggesting a joint role of features and structure. \textsf{NAEx}-4 shows a siginificant improvement suggesting the explanations are highly dependent on the structural similarities.

\subsection{Efficiency evaluation} 
We provide the training time of \textsf{NAEx} over $200$ random anchors for each dataset in Figure \ref{fig:eff}(a) and its comparative inference time with the baselines in Figure \ref{fig:eff}(b). As GNNExplainer and dNAME are not inductive and require parameter re-evaluation, they have the highest inference time. In comparison, Atten and Grad have a similar inference time as of \textsf{NAEx}, requiring either a single backward pass or none at all. However, Atten is model specifc and cannot be applied to any other NA model other than SANA. Once trained, \textsf{NAEx} can be used for inferencing in linear time, making it more practical for large-scale datasets.



\subsection{Qualitative evaluation}
We select an anchor from the ACM-DBLP dataset to illustrate the effectiveness of \textsf{NAEx} in identifying explanatory patterns. Figure 5 depicts the high-level representative explanations generated by \textsf{NAEx}. Thin black edges represent the original graph structure, red bold edges indicate the generated explanations, and pink nodes denote the extended substructures, which are nearby nodes to provide context of the larger neighborhood. The generated explanation offers several key insights: \textsf{NAEx} successfully identifies matching substructures (higher-order structural patterns, e.g., triadic closure in this case) across aligned node pairs. The generated explanations are both sparse, highlighting a minimal yet informative subset of edges, and also structurally coherent, meaning the selected edges form connected regions that span meaningful neighborhoods rather than isolated or disjoint fragments.

\noindent \underline{\textbf{NAEx for alignment model selection}}. \textsf{NAEx} can be used for task-specific model selection. One of \textsf{NAEx}’s strengths lies in its ability to generate separate masks for edges and node features, allowing a detailed decomposition of a model’s decision-making process as shown in ablation experiments. This can help select NA models suited to specific domains—for instance, in networks with sparse or noisy features but rich structural patterns (like social networks), a model with high edge-mask fidelity is preferable, whereas in attribute-rich settings (biological networks), feature-focused explanations may indicate stronger generalization. \textsf{NAEx} also helps pinpoint why an NA model may misalign node pairs. For example, if explanations frequently highlight irrelevant features or peripheral edges, this suggests the base NA model may be overfitting to noise or under-utilizing structure. This can guide debugging and architecture refinement.

\begin{figure}[t]
\centering
    \includegraphics[width=45mm,height=21mm]{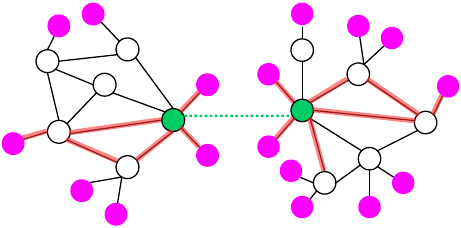}
\caption{A case anchor explanation on ACM-DBLP dataset}
\vspace{13pt}%
\end{figure}

\section{Conclusion}
In this paper, we proposed \textsf{NAEx}, a model-agnostic framework for explaining network alignment predictions. By leveraging a mutual information-based approach, \textsf{NAEx} identifies key subgraphs and features around anchor nodes in source and target networks, capturing their joint dependence through cross-graph interactions. This versatile framework is applicable across diverse domains, aiding in understanding and validating alignment models. Experiments on three benchmark datasets demonstrate its high fidelity in preserving alignment predictions using sparse explanatory subgraphs and features. Future work could extend \textsf{NAEx} to dynamic networks, enabling explanations in evolving graph scenarios and uncovering temporal patterns.





\bibliography{mybibfile}

\end{document}